\documentclass[conference]{IEEEtran}
\IEEEoverridecommandlockouts
\pdfoutput=1
\usepackage{cite}
\usepackage{amsmath,amssymb,amsfonts}
\usepackage{algorithmic}
\usepackage{graphicx}
\usepackage{textcomp}
\usepackage{xcolor}
\def\BibTeX{{\rm B\kern-.05em{\sc i\kern-.025em b}\kern-.08em
    T\kern-.1667em\lower.7ex\hbox{E}\kern-.125emX}}
\usepackage{float}
\usepackage[vlined,boxed,linesnumbered,commentsnumbered]{algorithm2e}

\newenvironment{alg}[1][htb]
               {% Update algorithm name
                 \begin{algorithm}[#1]%
               }{\end{algorithm}}

\usepackage{url}

\begin{document}

\title{Deep Active Learning with Budget Annotation}
\author{\IEEEauthorblockN{Kinyua Gikunda}
\IEEEauthorblockA{\textit{School of Computer Science and Information Technology} \\
\textit{Dedan Kimathi University of Technology}\\
Nyeri, Kenya \\
ORCID: 0000-0001-7962-2168}
}

\maketitle

\begin{abstract}
Digital data collected over the decades and data currently being produced with use of information technology is vastly the unlabeled data or data without description. The unlabeled data is relatively easy to acquire but expensive to label even with use of domain experts. Most of the recent works focus on use of active learning with uncertainty metrics measure to address this problem. Although most uncertainty selection strategies are very effective, they fail to take informativeness of the unlabeled instances into account and are prone to querying outliers. In order to address these challenges we propose an hybrid approach of computing both the uncertainty and informativeness of an instance, then automaticaly label the computed instances using budget annotator. To reduce the annotation cost, we employ the state-of-the-art pre-trained models in order to avoid querying information already contained in those models. Our extensive experiments on different sets of datasets demonstrate the efficacy of the proposed approach.
\end{abstract}


\section{Introduction}
Curent ICT technologies include Internet of Things(IoT) \cite{Weber_2010}, Remote Sensing(RS) \cite{Anindya_2016}, Cloud Computing(CC) \cite{Jinbo_2018} and Big Data(BD) \cite{Chi_2016}. RS entails detecting and monitoring the physical aspects of a target area from distance. CC offers tools for pre-processing and modelling of data collected from IoT devices and other sources. The continuous use of these technologies to collect, monitor, measure, store and analyse data has led to a phenomena of BD \cite{Chen_2014} which is in abundance of unlabeled data. Unlabeled data is relatively easy to acquire and it is expensive to label even with use of domain experts. For example, its expensive to hire dermatologists to annotate 129,450 skin cancer images \cite{Esteva_2017}.  Even when using state-of-the-art computing resources, training a Machine Learning(ML) model on large data sets can take long time. However, majority of the time ML algorithm may not need all of the available dataset for training \cite{Long_2016}. The main motivation for use of Active Learning(AL) is that, if a learning algorithm can pick the data it want to learn from, then a small set of selected datapoints will be used for training. Typically this process would involve randomly sampling large amount of data from underlying distribution for training a model(passive learning). Collecting large amount of labeled data for training is time consuming and expensive. AL provides methods for analyzing vast amount of data with improved efficiency than other computing approaches, because of the ability to iteratively select the most informative data sample and simultenously update its selection strategy \cite{Yarin_2017}. AL is a semi-supervised method that does not require labels of all the samples in a dataset. In unsupervised methods no labeled samples are used and for fully supervised all samples are labeled. The decision of how much data to use for training a Deep Learning Model or alternatively the level of performance required is a resource management decision.

In AL there are three scenarios in which the ML algorithms will query the labels of instances, they include; a) Membership Query Synthesis (MQS) - ML algorithms generates constructs of an instance \cite{Angluin_1987}. b) Stream-Based Selective Sampling (SSS) - ML algorithms use query strategy to determine whether to query the label of an instance or reject it based on informativeness \cite{Zhu_2007}. c) Pool-Based Sampling (PBS) - Instances are drawn from a pool of unlabeled data according to some informativeness measure \cite{Nigam_1998}. Majority of recent works focuses on use of pool-based sampling approach. In their work Joshi et al. (2009) proposes an uncertainty measure that generalizes margin based uncertainty to the multi-class \cite{Joshi_2009}. Chakraborty et al. (2011) propose a dynamic batch mode AL combined with selection criteria as a single formulation \cite{Chakraborty_2011}. In another recent study AL and Random Sampling(RS) is used to subtitute the human annotators \cite{Yang_2018}. The emphasis is to evaluate the informativeness of an instance, with an assumption that an instance with higher classification uncertainty is more crucial to label. This classical approach usually uses statistical theory such as entropy and margin to measure instance utility , however it fails to capture the data distributon information contained in the unlabeled data. This can eventually cause the classifier to select outlier instances to label therefore, its important to consider the classification uncertaninty as well as instance reprensentativeness while developing an AL method. 

The conventional way to reduce the cost of designing deep learning architectures and optimizing the parameters is to exploit available pre-trained models. Using pre-trained models helps reduce the training cost by utilizing information contained in different models but from a related domain. This is also refered to as Transfer Learning(TL)\cite{Sinno_2010}. The information transfer between the source and the target domain is done through feature sharing \cite{Eric_2014} and components transformation \cite{Sinno_2014}. However, these methods only revise the designs of the pre-trained networks, hence re-training with many parameters is still computing intensive. Performing batch training with fixed architecture of pre-trained models is a better choice . Classical state-of-the-art deep network architectures include: AlexNet \cite{Krizhevsky_2012}, NIN \cite{Lin_2013}, ENet \cite{Paszke_2016}, ZFNet \cite{Zeiler_2014}, GoogleLeNet \cite{Szegedy_2015} and VGG 16 \cite{Simonyan_2014}. Modern architectures include: Inception \cite{Szegedy_2016}, ResNet \cite{He_2015}, and DenseNet \cite{Huang_2017}. These networks have achieved impressive performance on computer vision, speech and text recognition  with effective representations for visual objects \cite{Gikunda_2019}. One way to exploit a pre-trained model is to use the entire network except the output layer as the feature extractor. However, when the target task is not similar to the source, the extracted feature is less effective for re-training \cite{Jason_2014}. A possible approach is to fine-tune the weight of the pre-trained model during target model training. Such methods can partially reduce the training cost, but still require a relatively large dataset to optimize the network weights \cite{Zhangjie_2017}. Neural networks presented in section 2 do not consider use of the pre-trained models, leading to waste of annotation cost since they query information already contained in pre-trained models.

In this paper, we purpose to perform a batch training using pre-trained state-of-the-art deep convolutional neural networks so as to reduce the annotation cost by not querying information already contained in the pretrained models. The data to label is selected by computing both the uncertainty metric and informativeness metric of an instance then budget labeling is implemented to automatically label the training data. The architecture is loaded with initialized parameters while performing high level visualization of accuracies and losses in real time. The selected instances are expected to be most useful for the classifier training, budget labeling and representation learning. We perform various experiments on batched SVHN, and CIFAR-10 datasets using modern architectures namely: Inception3, DenseNet and Squeezenet.

\section{Literature Review} \label{sec:literature}
Successful investigations on ways to reduce labeling cost by use of AL has been going on for years now \cite{Cohn_1994}. AL helps reduce the training data by selecting the most informative instances to query for labels \cite{Settles_2019}. In a typical AL model, learning proceeds sequentially, with the learning algorithm actively asking for the labels of some instances from a  membership queries(MQ). The intention is to query labels of the most informative instances, consequently reducing labeling costs and accelerating the learning. In recent time, there are a number of studies using AL strategies to reduce the training data. In \cite{Lin_2017} the authors explore ways to segment boiomedical images by combining fully convolutional network (FCN) and AL to reduce annotation effort by making suggestions on the most effective annotation areas. 

In their approach, FCN is used to provide uncertainty and similarity information which is used to evaluate the most informative areas for anotation. Sener et al. (2018) \cite{Ozan_2015} defines the problem of AL as a core-set selection by choosing a set of points that the model can use to learn in a batch setting environment. Wanh et al.(2017) \cite{Keze_2017} introduced a framework for updating the feature representation and the classifier simultaneously. A sample selection strategy is used to improve the classifier performance while reducing the manual annotation. 

The authors of\cite{Ozan_2015} transform AL into a core-set selection problem in batch setting, for selecting most competatve data points from the unlabeled set. A geometrical method is used to characterize the performance of the selected subset. The method proposed in \cite{Huang_2018} use fine tuned pre-trained model on most useful examples. The examples are estimated based on potential contribution of an instance to feature representation. Iscen et al (2019) \cite{Iscen_2019} introduce a transductive method that uses nearest neighbour graph to make predictions for generating pseudo-labels of the unlabeled dataset. Other studies considers queries for TL with classical shallow models. For example, the method in \cite{Xuezhi_2014} combines AL and TL into a Gaussian process based approach, and sequentially uses predictive covariance to select most suitable queries from the target domain. 

Kale and Liu \cite{David_2013} propose a framework to combine the AL with TL, and utilize labeled data from source domain to improve the performance in the target domain. Kale et al. \cite{David_2015}  present a framework for generating effective label queries by performing TL. The framework is able to perform both the un-supervised and semi-supervised learning. Huang and Chen (2016) \cite{Sheng_2016}  propose to actively query labels from source domain to help the learning task of the target domain. From the literature presented above, majority of the AL focus on selecting a single informative unlabeled instance to label each time. One main shortcoming of the above approaches is poor generalization for unseen instances in the domain. This is due to the fact that they only select queries based on how the instance related to the classsifier while ignoring unlabeled instances. Also with a large set of instances classification response time can be slow, therefore use of budget anotator will help reduce active labeleling evaluation metrics and labeling time.

\section{Methodology} \label{sec:method}

\begin{figure*}[ht]
 \centering
  \includegraphics[width=.7\textwidth]{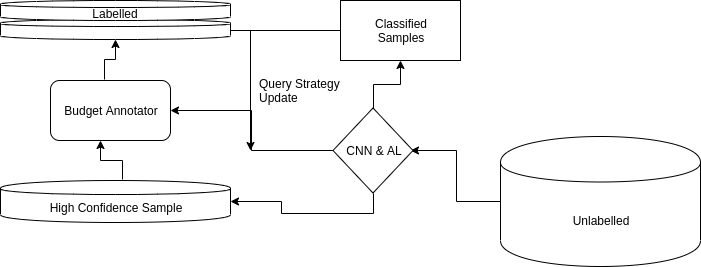}
 \caption{\small{Our proposed model. AL framework progressively get data as input from the unlabeled set. Most informative samples and the classified samples are applied on the classifier output. The process to select and label instances will iterate until the budget is achived while updating the selection strategy}}
 	\label{fig:al}
 \end{figure*}

In order to avoid the problem of generalization of unseen instances, we present a robust approach by combining the strenghts of different learning strategies. The proposed approach has four main components: a) an uncertainty measure, b) correlation measure, c) an informative measure and d) and budget labeller.

We will use the follow notation in this paper. While $D = D^L \cup D^U$ let $D^L$ denote labelled instances $D^L = \{(x_1,y_1),(x_2,y_2)....(x_n,y_n)\}$, $D^U$ denote unlabelled instances $D^U = \{(x_1,?),(x_2,?)....(x_n,?)\}$, $D^H$ denote high confidence instances and $\theta^L$ denote the model parameters. For $m$ classes in $D$ the label of $D^U$ can be expressed as $y^i = l, l\in \{1,2,...,m\}$. Therefore, instance selection creteria in this work will be based on probability of $x_i$ belonging to $l$th class which can be expressed as $p(y_i = l|x_i;\theta)$. The algorithm b describes the general axctive learning algorithm. We follow the same approach to develop the proposed methodology.

\begin{alg}[ht]
\SetKwData{Left}{left}\SetKwData{This}{this}\SetKwData{Up}{up}
\SetKwFunction{Union}{Union}\SetKwFunction{FindCompress}{FindCompress}
\SetKwInOut{Input}{input}\SetKwInOut{Output}{output}
%\Input{$\mathcal{P}$ the current board position}
%\Output{the best move}
%\BlankLine
    {\bf Require:} labelled instance set $D^L$, unlabelled instance set $D^U$, size of the training set $m$ \;
    {\bf Ensure:} model $\Theta$\;
\While{training size $ \leqslant m $}{
  $\Theta \leftarrow $ learn a model based on $D^L$\;
  $D^U \leftarrow D \setminus D^L $\;
  \For{each $x_i$ in $D^U$} {
  $u_i \leftarrow u(x_i, \Theta)$\;}
  $x^* \leftarrow \underset{i}{\mathrm{argmax}}(u_i)$\;
  $D^l \leftarrow D^L \cup \{x^*\}$ \;
% $D^U \leftarrow D^U \setminus x^*$ \;  
}
update $\Theta$ based on $D^L$\;
\label{alg:alg2}
\caption{General AL algorithm adopted from \cite{Fu_2012} We follow the same approach to develop the AL with budget annotation, however, in their approach step number 3 not necessary since the unlabelled set has been updated in step number 9}
\end{alg}

From Algorithm \ref{alg:alg2}. a model is trained from a small set $D^L$. Then instances from unlabelled pool $D^U$ are queried based on evaluated uncertainty measure and instance informativeness. As the learning process continue, the model will sequentially update its selection strategy while selecting more high confidence samples. 

\section{Experiments} \label{sec:experiment}

\subsection{Datasets} \label{subsec:datasets}
Two datasets namely CIFAR-10 \cite{cifar10} and Street View House Numbers (SVHN) \cite{svhn} datasets were used in this work. The CIFAR-10 dataset used consist of 32x32 10,000 labelled image pool, 30,000 image unlabelled pool and 10,000 testing pool. SVHN dataset used consist of 32x32 10,000 labelled image pool, 30,000 unlabelled pool and 26,032 testing pool. For both datasets, input resize and normalize transformation was done in order to match the models input sizes and shapes. The number of initial labels was initialized to 10,000 and number of queries to 1,000. 15 training epochs were carried out with a training batch size of 32 and and a learning rate of 0.05. The size of batch size was considered in order to fit the entire training batch into memory since GPU was used for speed up \cite{Bengio_2012}. This also ensure achieving a good training stability and generalization. The proposed approach was implemented in Python 3.6 .Nvida Tesla P100 was for experiments. During all the experiments, losses and accuracies set were monitored. 

\subsection{Fine-tuning Network Parameters } \label{subsec:np}
In order to suite the pre-trained network to the dataset classes, the last layer(softmax layer) is truncated and replaced with a layer with 10 categories. Back propagation is performed to fine-tune the pre-trained weights. Since the pre-trained networks are already good as compared to initialized weights, we adjsut the learning rate to 0.05. 

\subsection{Models} \label{subsec:models}
In the experiments, the study was done using the following three state-of-the-art pre-trained models which have achieved a top-5 error rate in ILSVRC. In this section we will briefly discuss the architectures of the selected models. 

GoogleNet, a 2014 ILSVRC winner, was inspired by LeNet but implemented a novel inception module. he Inception cell performs series of convolutions at different scales and subsequently aggregate the results. This module is based on several very small convolutions in order to drastically reduce the number of parameters. There has been tremedious efforts done to improve the performance of the architecture: a) Inception v1 \cite{Szegedy_2015} which performs convolution on an input, with 3 different sizes of filters (1x1, 3x3, 5x5). Additionally, max pooling is also performed. The outputs are concatenated and sent to the next inception module. b) Inception v2 and Inception v3 factorize 5x5 convolution to two 3x3 convolution operations to improve computational speed. Although this may seem counterintuitive, a 5x5 convolution is 2.78 times more expensive than a 3x3 convolution. So stacking two 3x3 convolutions infact leads to a boost in performance. c) In Inception v4 and Inception-ResNet the initial set of operations were modified before introducing the Inception blocks.

\begin{figure*}[ht]
\minipage{1.0\textwidth}
  \includegraphics[width=\linewidth]{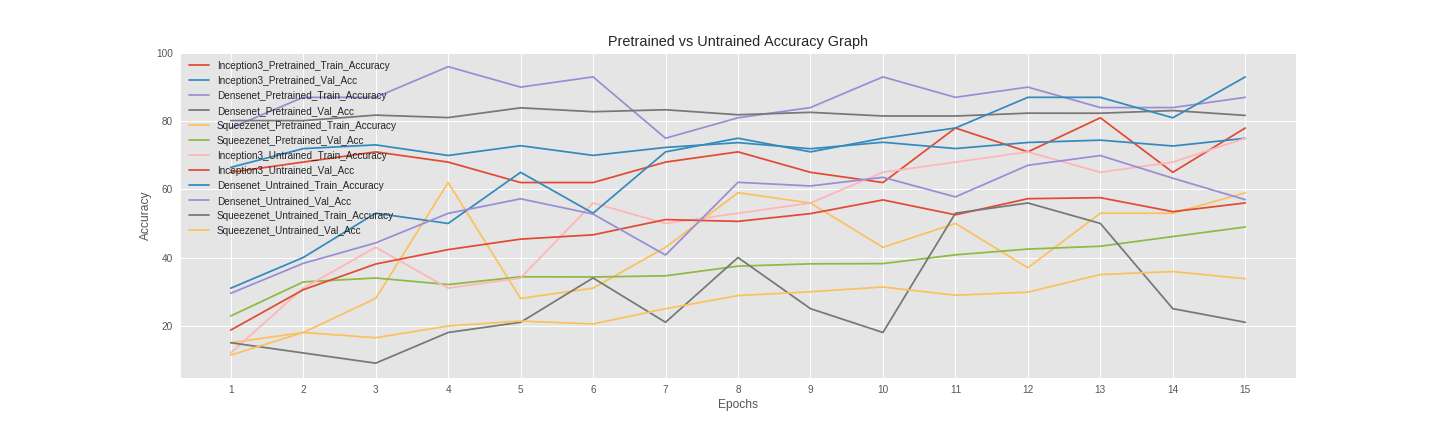}
  \caption{Scratch Vs Pre-trained DenseNet }\label{fig:scratch_pretrained}
\endminipage\hfill
\minipage{1.0\textwidth}
  \includegraphics[width=\linewidth]{pretrained_untrained_acc.png}
  \caption{Pre-trained DenseNet with AL on SVHN dataset. }\label{fig:densenet}
\endminipage\hfill
\end{figure*}

\begin{figure*}[ht]
\minipage{1\textwidth}
  \includegraphics[width=\linewidth]{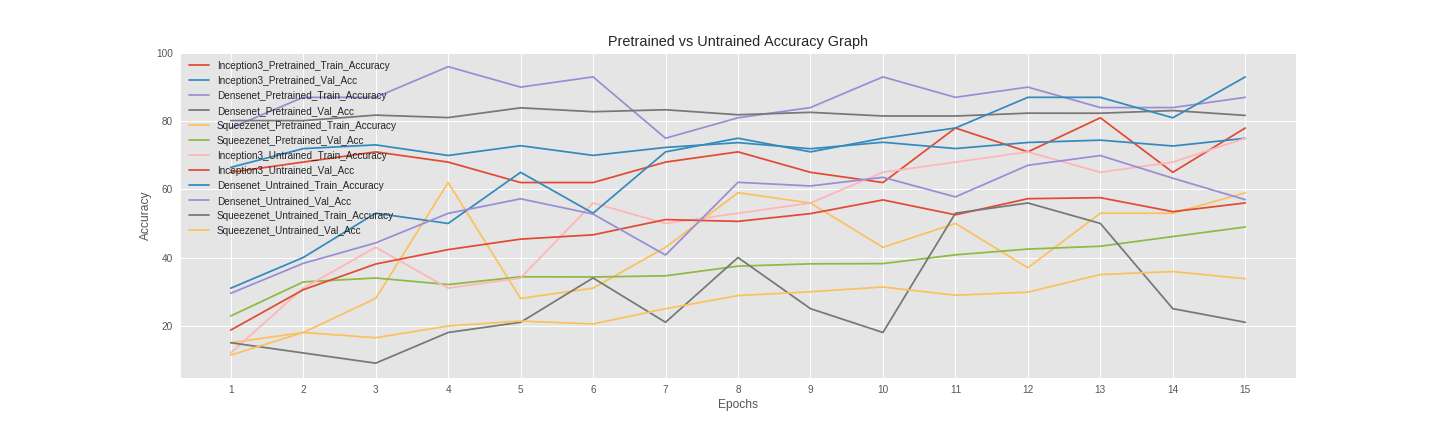}
  \caption{ResNet AL and auto labeler on batched SVHN dataset  }\label{fig:resnet}
\endminipage\hfill
\minipage{1\textwidth}
  \includegraphics[width=\linewidth]{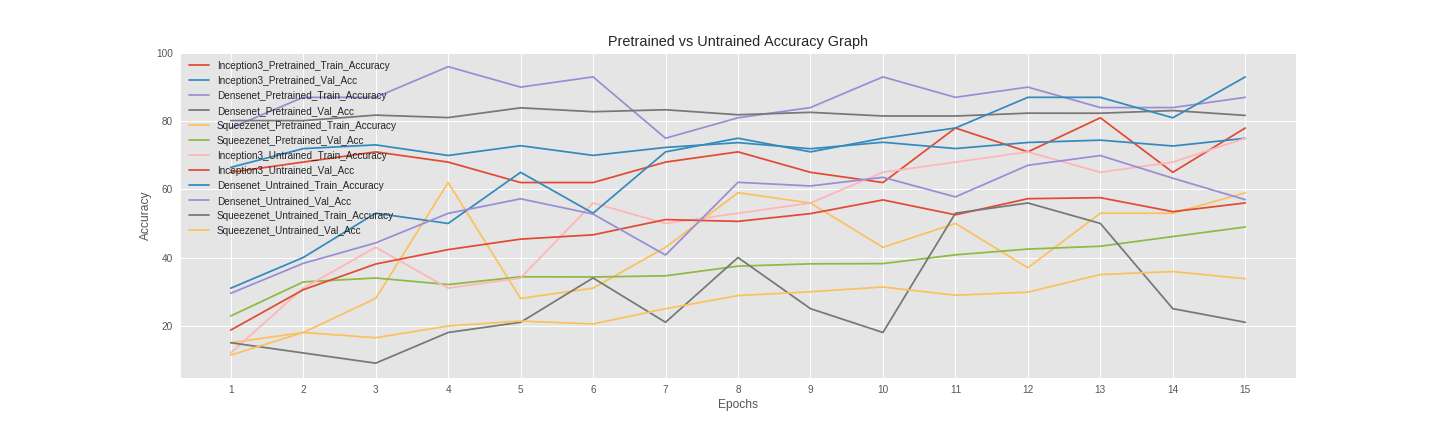}
  \caption{GoogleNet AL and auto anotator on batched MNIST dataset}\label{fig:googlenet}
\endminipage\hfill
\end{figure*}

Extensive classification study of AL and budget anotator with DenseNet, GoogleNet and ResNet  on batched SVHN sataset in comparison to classical active learning without budget labelling. Our proposed method AL with budget labelling performs consistently better than classical AL method. Fig. \ref{fig:scratch_pretrained}. indicate impressive prediction performance between using pre-trained models and training from scratch models.  Fig. \ref{fig:densenet}. shows pre-trained DenseNet with performance of up to 92\% validation accuracy on SVHN dataset(a pool of 40,000 images was used. 10,000 labelled set, 30,000 unlabelled set and 10,000 validation set). This impressive performance imply that use of both auto anotators and pe-trained models can be used to train a deep CNN using a small labelled dataset over a big pool of unlabelled data. In Fig. \ref{fig:resnet}. we trained ResNet on  AL and pseudo anotator with batched SVHN dataset.In Fig. \ref{fig:googlenet}. we show the applicability of our approach on different datasets, we used a pre-trained models on MNIST dataset. 

When deeper networks starts converging, a degradation problem is exposed, with the network depth increasing, accuracy gets saturated and then degrades rapidly. Deep Residual Neural Network(ResNet), a logical extension of DenseNet \cite{Huang_2017} created by Kaiming He al. \cite{He_2015}introduced a norvel architecture with insert shortcut connections which turn the network into its counterpart residual version. This was a breakthrough which enabled the development of much deeper networks. The residual function is a step in which the network learn how to adjust the input feature map for higher quality features. Following this intuition, the network residual block authors proposed a pre-activation variant of residual block, in which the gradients can flow through the shortcut connections to any other earlier layer unimpeded. Each ResNet block is either 2 layer deep (used in small networks like ResNet 18, 34) or 3 layer deep(ResNet 50, 101, 152). It achieves a top-5 error rate of 3.57\% which beats human-level performance on this dataset.

DenseNet which is a logical extension of ResNet, there is improved efficiency by concatenating each layer feature map to every successive layer within a dense block \cite{Huang_2017}. This allows later layers within the network to directly leverage the features from earlier layers, encouraging feature reuse within the network. For each layer, the feature-maps of all preceding layers are used as inputs, and its own feature-maps are used as inputs into all subsequent layers, this helps alleviate the vanishing-gradient problem, feature reuse and reduce number of parameters.

\section{Conclusion} \label{sec:conclusion}
In this paper, we propose an budget anotator AL approach for cost-effective training of deep convolutional neural networks. Instead of training from scratch, a pre-trained model can be effectively adapted to a new target task by fine tuning with a few actively queried examples, significantly reducing the cost of designing the network architecture and labelling a large training set. Using AL with budget labelling, one can achieve up to 90\% prediction accuracy with little amount of training data as compared to conventional training. The use of budget labelling techique ensures that the model automatically updates its selection strategy after every iteration.

\vspace{12pt}


\begin{thebibliography}{100}

\bibitem{Weber_2010}
Weber R. H. Weber R. Internet of Things: Legal perspectives. Berlin: Springer-Verlag Berlin Heidelberg (2010).10.1007/978-3-642-11710-7 

\bibitem{Anindya_2016}
Anindya Sunday, Remote Sensing in Agriculture, International Journal of Environment, Agriculture and Biotechnology (IJEAB)  Vol-1, Issue-3 (2016)

\bibitem{Jinbo_2018}
Jinbo, C., Xiangliang, C., Han-Chi, F. et al. Cluster Computing (2018).\url{ https://doi.org/10.1007/s10586-018-2022-5}

\bibitem{Chi_2016}
Chi M, Plaza A, Benediktsson JA, Sun Z, Shen J and Zhu Y. Big data for remote sensing: challenges and opportunities. Proceedings of the IEEE 104, 2207–2219 (2016)

\bibitem{Chen_2014}
Chen, M., Mao, S.,  Liu, Y. Big Data: A Survey. MONET, 19, 171-209 (2014)

\bibitem{Esteva_2017}
Esteva, Andre, Kuprel, Brett, Novoa, Roberto A, Ko, Justin, Swetter, Susan M, Blau, Helen M, \& Thrun, Sebastian. Dermatologist-level classification of skin cancer with deep neural networks. Nature, 542(7639):115–118 (2017)

\bibitem{Long_2016}
Long, M., Zhu, H., Wang, J., \& Jordan, M.I. Deep Transfer Learning with Joint Adaptation Networks. ICML (2016)

\bibitem{Yarin_2017}
Yarin Gal, Riashat Islam, and Zoubin Ghahramani. Deep Bayesian Active Learning with Image Data. In International Conference on Machine Learning. 1183–1192

\bibitem{Angluin_1987}
Angluin, D. Queries and concept learning. Machine Learning, 2, 319-342 (1987)

\bibitem{Zhu_2007}
Zhu, X., Zhang, P., Lin, X., \& Shi, Y. Active Learning from Data Streams. Seventh IEEE International Conference on Data Mining (ICDM 2007), 757-762 (2007)

\bibitem{Nigam_1998}
Nigam, K., \& McCallum, A. Pool-Based Active Learning for Text Classification (1998)

\bibitem{Joshi_2009} 
Joshi, A.J., Porikli, F.M., \& Papanikolopoulos, N. Multi-class active learning for image classification. CVPR (2009)

\bibitem{Chakraborty_2011}
Chakraborty, S., Balasubramanian, V.N., \& Panchanathan, S). Dynamic batch mode active learning. CVPR 2011, 2649-2656 (2011)

\bibitem{Yang_2018}
Yang, Y., \& Loog, M. Single shot active learning using pseudo annotators. Pattern Recognition, 89, 22-31 (2018)

\bibitem{Sinno_2010}
Sinno Jialin Pan and Qiang Yang. 2010. A Survey on Transfer Learning. IEEE Transactions on Knowledge and Data Engineering 22, 10 , 1345–1359 (2010)

\bibitem{Eric_2014}
Eric Tzeng, Judy Hoffman, Ning Zhang, Kate Saenko, and Trevor Darrell. 2014. Deep Domain Confusion: Maximizing for Domain Invariance. CoRR abs/1412.3474. arXiv:1412.3474 (2014)

\bibitem{Sinno_2014}
Sinno Jialin Pan, Ivor W Tsang, James T Kwok, and Qiang Yang. 2011. Domain adaptation via transfer component analysis. IEEE Transactions on Neural Networks 22, 2, 199–210 (2011)

\bibitem{Gikunda_2019}
Gikunda P.K., Jouandeau N.State-of-the-Art Convolutional Neural Networks for Smart Farms:  Advances in Intelligent Systems and Computing, vol 997. Springer, (2019) 

\bibitem{Krizhevsky_2012}
Krizhevsky, A., Sutskever, I., Hinton, G.E.: ImageNet classification with deep convolutional neural networks. Commun. ACM 60, 84–90 (2012)

\bibitem{Lin_2013}
Lin, M., Chen, Q., Yan, S.: Network in network. CoRR, abs/1312.4400 (2013)

\bibitem{Paszke_2016}
Paszke, A., Chaurasia, A., Kim, S., Culurciello, E.: ENet: a deep neural network architecture for real-time semantic segmentation. CoRR, abs/1606.02147 (2016)

\bibitem{Zeiler_2014}
Zeiler, M.D., Fergus, R.: Visualizing and understanding convolutional networks. In: European Conference on Computer Vision, pp. 818–833. Springer (2014)

\bibitem {svhn}
The Street View House Numbers (SVHN) Dataset \url{http://ufldl.stanford.edu/housenumbers/}

\bibitem {cifar10}
CIFAR-10 dataset \url{https://www.cs.toronto.edu/~kriz/cifar.html}

\bibitem{Bengio_2012}
Bengio, Y. Practical recommendations for gradient-based training of deep architectures. Neural Networks: Tricks of the Trade (2012)

\bibitem{Szegedy_2015}
Szegedy, C., Liu, W., Jia, Y., Sermanet, P., Reed, S.E., Anguelov, D., Erhan, D., Vanhoucke, V., Rabinovich, A.: Going deeper with convolutions. In: IEEE Conference on Computer Vision and Pattern Recognition (CVPR), pp. 1–9 (2015)

\bibitem{Simonyan_2014}
Simonyan, K., Zisserman, A.: Very deep convolutional networks for large-scale image recognition. CoRR, abs/1409.1556 (2014)

\bibitem{Szegedy_2016}
Szegedy, C., Vanhoucke, V., Ioffe, S., Shlens, J., Wojna, Z.: Rethinking the inception architecture for computer vision. In: IEEE Conference on Computer Vision and Pattern Recognition (CVPR), pp. 2818–2826 (2016)

\bibitem{He_2015}
He, K., Zhang, X., Ren, S., Sun, J.: Deep residual learning for image recognition. In: 2016 IEEE Conference on Computer Vision and Pattern Recognition (CVPR), pp. 770–778 (2015)

\bibitem{Huang_2017}
Huang, G., Liu, Z., Maaten, L.V., Weinberger, K.Q.: Densely connected convolutional networks. In: 2017 IEEE Conference on Computer Vision and Pattern Recognition (CVPR), pp. 2261–2269 (2017)

\bibitem{Jason_2014}
Jason Yosinski, Jeff Clune, Yoshua Bengio, and Hod Lipson. How transferable are features in deep neural networks?. In Advances in Neural Information Processing Systems 27. 3320–3328 (2014).

\bibitem{Zhangjie_2017}
Zhangjie Cao, Mingsheng Long, Jianmin Wang, and Michael I. Jordan. Partial Transfer Learning with Selective Adversarial Networks. CoRR abs/1707.07901. arXiv:1707.07901, (2017)

\bibitem{Keze_2017}
Keze Wang, Dongyu Zhang, Ya Li, Ruimao Zhang, and Liang Lin. Cost-Effective Active Learning for Deep Image Classification. IEEE Transactions on Circuits and Systems for Video Technology 27, 12 (2017), 2591–2600, 2017

\bibitem{Huang_2018}
Huang, S., Zhao, J., \& Liu, Z. Cost-Effective Training of Deep CNNs with Active Model Adaptation. KDD (2018)

\bibitem{Iscen_2019}
Iscen, A., Tolias, G., Avrithis, Y., \& Chum, O. Label Propagation for Deep Semi-supervised Learning. ArXiv, abs/1904.04717 (2019)

\bibitem{Lin_2017}
Lin Yang, Yizhe Zhang, Jianxu Chen, Siyuan Zhang, and Danny Z. Chen. Suggestive Annotation: A Deep Active Learning Framework for Biomedical Image Segmentation. In Medical Image Computing and Computer Assisted Intervention  399–407 (2017)

\bibitem{Fu_2012}
Fu, Y., Zhu, X., \& Li, B. A survey on instance selection for active learning. Knowledge and Information Systems, 35, 249-283 (2012)

\bibitem{Cohn_1994}
Cohn, D., Atlas, L. \& Ladner, R. Mach Learn 15: 201. https://doi.org/10.1007/BF00993277 (1994).

\bibitem{Ozan_2015}
Ozan Sener and Silvio Savarese. Active Learning for Convolutional Neural Networks: A Core-Set Approach. stat 1050, 27  (2017).

\bibitem{Settles_2019}
B. Settles, “Active learning literature survey,” Comput. Sci. Dept., Univ. Wisconsin–Madison, Madison, WI, USA, Tech. Rep. 1648, (2009)

\bibitem{Xuezhi_2014} 
Xuezhi Wang, Tzu-Kuo Huang, and Jeff Schneider. Active transfer learning under model shift. In International Conference on Machine Learning. 1305–1313. 399–407, 2014.

\bibitem{David_2013} 
David Kale and Yan Liu. Accelerating active learning with transfer learning. In IEEE 13th International Conference on Data Mining. 1085–1090 (2013)

\bibitem{David_2015} 
David C. Kale, Marjan Ghazvininejad, Anil Ramakrishna, Jingrui He, and Yan Liu. Hierarchical active transfer learning. In The SIAM International Conference on Data Mining. 514–522 (2015)

\bibitem{Sheng_2016} 
Sheng-Jun Huang and Songcan Chen. Transfer learning with active queries from source domain. In The 25th International Joint Conference on Artificial Intelligence. 1592–1598 (2016)

\bibitem{Szummer_2002}
Szummer, M., \& Jaakkola, T.S. Information Regularization with Partially labeled Data. NIPS (2002)

\bibitem{Huang_2017}
Huang, G., Liu, Z., Maaten, L.V., Weinberger, K.Q. Densely Connected Convolutional Networks. 2017 IEEE Conference on Computer Vision and Pattern Recognition (CVPR), 2261-2269 (2017)

\end{thebibliography}
\end{document}